\pdfoutput=1
\documentclass[11pt]{article}
\usepackage[a4paper,margin=2.5cm]{geometry}
\usepackage[T1]{fontenc}
\usepackage{lmodern}
\usepackage{amsmath,amssymb}
\usepackage{graphicx}
\usepackage{booktabs}
\usepackage{tabularx}
\usepackage{caption}
\usepackage[hidelinks]{hyperref}
\usepackage{enumitem}
\usepackage{microtype}
\usepackage{textcomp}
\makeatletter
\begingroup
\catcode`'=\active
\catcode``=\active
\g@addto@macro\@noligs{\let'\textquotesingle\let`\textasciigrave}
\endgroup
\makeatother
\setlist{nosep,leftmargin=*}
\newcommand{\astra}{GPT-6 Astra}
\newcommand{\istN}{19{,}435}
\newcommand{\istCountries}{36}
\newcommand{\nTrain}{11{,}749}
\newcommand{\nTest}{7{,}686}
\newcommand{\trainFrom}{January 1991}
\newcommand{\trainTo}{March 1995}
\newcommand{\testFrom}{April 1995}
\newcommand{\testTo}{May 1996}
\newcommand{\modelCLow}{0.789}
\newcommand{\modelCHigh}{0.794}
\newcommand{\nTestDD}{7{,}578}
\newcommand{\rateDD}{61.5}

\newcommand{\rateSix}{21.5}
\newcommand{\nTrainDD}{11{,}707}
\newcommand{\runsBlack}{96}
\newcommand{\runsWhite}{6}
\newcommand{\runsAll}{102}

\newcommand{\runMinutes}{16}
\newcommand{\runMinutesMax}{32}
\newcommand{\callsMedian}{12}
\newcommand{\callsMin}{7}
\newcommand{\callsMax}{21}
\newcommand{\callsBlack}{1{,}219}
\newcommand{\callsRefused}{15}
\newcommand{\callsFailed}{12}

\newcommand{\callsInsideWhite}{4}
\newcommand{\onlyCalculator}{every}

\newcommand{\rTwoLlm}{0.976}
\newcommand{\rTwoMedian}{0.982}
\newcommand{\rTwoAnalyst}{0.959}
\newcommand{\rTwoDiff}{+0.017}
\newcommand{\rTwoDiffCi}{0.016 to 0.018}
\newcommand{\rTwoDiffByModelCi}{0.012 to 0.022}
\newcommand{\rTwoLowest}{0.923}
\newcommand{\rTwoWhite}{0.951}
\newcommand{\rTwoBetter}{28}
\newcommand{\modelsAll}{32}
\newcommand{\qOneHolds}{met}
\newcommand{\rTwoWhiteBlack}{0.947}
\newcommand{\madLlm}{0.126}
\newcommand{\madAnalyst}{0.186}
\newcommand{\madBetter}{29}
\newcommand{\madLlmMlp}{0.269}
\newcommand{\termsMedian}{128}
\newcommand{\termsQOne}{115}
\newcommand{\termsQThree}{144}
\newcommand{\termsMin}{79}
\newcommand{\termsMax}{347}
\newcommand{\interMedian}{25.5}
\newcommand{\analystTerms}{42}
\newcommand{\foundK}{69}
\newcommand{\foundN}{72}
\newcommand{\foundPct}{96}
\newcommand{\foundCi}{80\% to 99\%}
\newcommand{\foundIcc}{1.00}
\newcommand{\foundNeff}{24}
\newcommand{\flawedModels}{24}
\newcommand{\foundAllThree}{23}
\newcommand{\foundNone}{1}

\newcommand{\qTwoHolds}{met}
\newcommand{\scriptFound}{24}
\newcommand{\scriptReports}{45}

\newcommand{\missedInput}{\texttt{visuospatial\_disorder}}
\newcommand{\missedKind}{logistic regression}
\newcommand{\missedOutcome}{death}
\newcommand{\missedEffect}{0.29}
\newcommand{\reverseNext}{0.44}
\newcommand{\reverseMax}{1.75}
\newcommand{\subgroupShareLow}{6}
\newcommand{\subgroupShareHigh}{13}
\newcommand{\reportsAll}{1{,}132}
\newcommand{\reportsVerifiable}{1{,}089}

\newcommand{\reportsFailed}{655}
\newcommand{\reportsFailedPct}{60}
\newcommand{\reportsFailedCi}{53\% to 67\%}

\newcommand{\classSmaller}{412}
\newcommand{\classSome}{79}
\newcommand{\classNotFound}{164}
\newcommand{\classSmallerPct}{63}
\newcommand{\classSomePct}{12}
\newcommand{\classNotFoundPct}{25}
\newcommand{\notFoundOfVerifiablePct}{15}
\newcommand{\perRunClean}{17.3}
\newcommand{\perRunFlawed}{10.0}
\newcommand{\perRunLr}{1.5}
\newcommand{\perRunEbm}{22.5}
\newcommand{\perRunHgb}{19.0}
\newcommand{\perRunMlp}{4.1}
\newcommand{\stepSharePct}{81}
\newcommand{\firstListed}{68}
\newcommand{\foundRuns}{69}
\newcommand{\problemsFoundRuns}{4}
\newcommand{\cleanRuns}{24}
\newcommand{\cleanWithProblem}{21}
\newcommand{\cleanFirstNotFound}{11}
\newcommand{\cleanFirstCorrect}{6}
\newcommand{\relLlmRawK}{68}
\newcommand{\relLlmRawN}{601}
\newcommand{\relLlmRawPct}{11}
\newcommand{\relLlmRawCi}{9\% to 14\%}
\newcommand{\relLlmCheckedK}{34}
\newcommand{\relLlmCheckedN}{103}
\newcommand{\relLlmCheckedPct}{33}
\newcommand{\relLlmCheckedCi}{25\% to 43\%}
\newcommand{\relScanRawK}{28}
\newcommand{\relScanRawN}{101}
\newcommand{\relScanRawPct}{28}
\newcommand{\relScanRawCi}{20\% to 37\%}
\newcommand{\relScanCheckedK}{25}
\newcommand{\relScanCheckedN}{82}
\newcommand{\relScanCheckedPct}{30}
\newcommand{\relScanCheckedCi}{22\% to 41\%}
\newcommand{\relRandomK}{20}
\newcommand{\relRandomN}{348}
\newcommand{\relRandomPct}{6}
\newcommand{\relRandomCi}{4\% to 9\%}

\newcommand{\relKnownN}{17}
\newcommand{\relDiff}{+3}
\newcommand{\relDiffCi}{$-$11 to 16}
\newcommand{\foundOrSameK}{69}
\newcommand{\candLlm}{68}
\newcommand{\candScan}{28}
\newcommand{\verLlm}{19}
\newcommand{\verScan}{22}
\newcommand{\verBoth}{16}
\newcommand{\verLlmOnly}{3}

\newcommand{\probC}{0.66}
\newcommand{\probBrier}{0.30}
\newcommand{\smallerSteps}{409}
\newcommand{\smallerStepsHgb}{245}
\newcommand{\smallerStepsEbm}{164}
\newcommand{\smallerStepMedian}{0.15}
\newcommand{\smallerStepUnderPct}{12.5}
\newcommand{\smallerOther}{3}
\newcommand{\distLlmK}{19}
\newcommand{\distLlmN}{66}
\newcommand{\distLlmPct}{29}
\newcommand{\distScanK}{22}
\newcommand{\distScanN}{79}
\newcommand{\distScanPct}{28}
\newcommand{\hProposed}{348}
\newcommand{\hHoldK}{35}
\newcommand{\hHoldPct}{10}

\newcommand{\hypPerRun}{58}
\newcommand{\hypMin}{12}
\newcommand{\hypMax}{155}

\hypersetup{pdftitle={Auditing black-box clinical prediction models with a large language model through queries alone: a blinded test with planted flaws},pdfauthor={Zhiyu Liu, Xinyu Yang}}

\title{Auditing black-box clinical prediction models with a large language model through queries alone: a blinded test with planted flaws}
\author{Zhiyu Liu\textsuperscript{1} \qquad Xinyu Yang\textsuperscript{2,*}\\[6pt]
{\small \textsuperscript{1}Tianjin Medical University, Tianjin, China; \href{mailto:07eb1f@gmail.com}{07eb1f@gmail.com}}\\
{\small \textsuperscript{2}Tianjin Medical University General Hospital, Tianjin, China; \href{mailto:yangxinyu@tmu.edu.cn}{yangxinyu@tmu.edu.cn}}\\
{\small \textsuperscript{*}Corresponding author}}
\date{November 2026}

\begin{document}
\maketitle

\begin{abstract}
\noindent Large language models (LLMs) are used to explain clinical prediction models, on the promise that they can say what a model computes, find its faults and even reveal new knowledge in it. We tested the three promises where answers could be checked. From the International Stroke Trial we trained four kinds of model (logistic regression, explainable boosting machine, gradient-boosted trees, neural network) for two outcomes on patients randomised earlier, planted one flaw in each of \flawedModels{} copies (an under-predicted subgroup, a reversed risk factor or an implausible step) and mixed them with the 8 clean models. \astra{} audited each model three times, blind, through a tool that only returned the model's predictions; the judging rules were committed before any run and applied by program. Its surrogate models, with a median of \termsMedian{} terms, reproduced the audited models' log-odds with a mean $R^2$ of \rTwoLlm{} (\rTwoAnalyst{} for an analyst's additive surrogate of \analystTerms{} terms). It found the planted flaw in \foundK{} of \foundN{} runs (\foundPct\%; 95\% CI \foundCi{}); a script written for the purpose found all \scriptFound{}. It also reported many more problems: \reportsFailedPct\% of \reportsVerifiable{} checkable reports failed the pre-specified check against the model's average behaviour; post hoc, most were present in the model but under the threshold, nearly all small steps of tree-based models (median \smallerStepMedian{} log-odds, no lower bound), and \notFoundOfVerifiablePct\% of the checkable reports were not found in the model. Of \relLlmRawN{} unexpected relationships it read out of the clean models, \relLlmRawPct\% held in patients randomised later (random pairs \relRandomPct\%; a regression scan's shorter list, \relScanRawPct\%); after a check on the training data, \relLlmCheckedPct\% against \relScanCheckedPct\% (\distLlmPct\% against \distScanPct\% for distinct relationships). With query access alone, an LLM can describe a black-box model faithfully and find nearly every planted flaw, as a purpose-written script also did; its reports of problems must be checked on the model and its discoveries on data.
\end{abstract}

\section{Introduction}

A clinician or a regulator who faces a prediction model wants to know three things: what it computes, where it is wrong, and whether the unexpected patterns it shows are real. For most models these questions are hard to answer. The model is a black box, post hoc explanations can mislead \cite{rudin2019,ghassemi2021}, and a commercial model is often available only through an interface that returns predictions.

Large language models (LLMs) have been proposed as a way through. They can read transparent additive models and point out surprising effects \cite{lengerich2023,bordt2024}, and it has been claimed that they can read and edit the weights of neural networks directly \cite{takehi2026}. In an earlier study with simulated networks whose content was known, we found that an LLM with a calculation tool described and audited such networks equally well whether it saw the weights or could only query the network: the ability came from computation, not from access to weights \cite{liu2026}. A similar audit, from weights, has been reported for rule networks in radiotherapy planning \cite{nusrat2026}.

What is missing is a test on models trained on real patient data in which the answers are known in advance. Real models carry natural irregularities and noise, so a planted error is harder to tell apart, and an auditor that reports everything it sees is not useful. Planting known defects and asking whether they are found is an established way of testing explanation methods and auditors in machine learning \cite{adebayo2020,marks2025}, and functions with known descriptions serve to test automated interpretability \cite{schwettmann2023,shaham2024}. And a faithful description of a model is not yet knowledge about patients: a model may have learned noise, and a faithful reader will report the noise.

We tested an LLM as an auditor of clinical prediction models under these conditions. Our contributions are:
\begin{itemize}
\item a protocol for blinded audits of prediction models trained on real data: flaws of three kinds are planted in the model's output, the auditor sees only predictions, it reports in a form that a program can check, and the rules for judging are committed before any run;
\item comparators that use no LLM for each question: an analyst's surrogate, a purpose-written audit script, and candidates from regression scans and from the model's own strongest interactions, with positive and negative controls;
\item a guard for the blinding: the tool shows nothing of the planted layer, even when a call fails, and every tool call was read afterwards;
\item results for \modelsAll{} models trained on an open trial database: faithful surrogates, planted flaws found in \foundPct\% of runs, many reported problems that do not pass a check on the model, and unexpected relationships that hold in later patients no more often than those from a regression scan;
\item all prompts, replies, tool calls, models and judging code.
\end{itemize}

\section{Related work}

\paragraph{LLMs that explain models.} TalkToModel lets a user question a tabular model in natural language, each question being translated into explanation operations run on the model \cite{slack2023}. TalkToEBM gives an LLM the graphs of an explainable boosting machine and reports that it describes them, finds surprises and suggests repairs \cite{lengerich2023,bordt2024}. Whether the surprises hold in new patients was not tested. We test this for relationships read out of four kinds of model, against candidates that need no LLM.

\paragraph{Reading weights.} Reports that frontier LLMs read and write neural network weights \cite{takehi2026}, and models trained to describe weights \cite{kabra2026}, raised the question of what such reading rests on; one study found that zero-shot reading of small image networks rested on local arithmetic and on familiar datasets rather than on tracing the network \cite{ahmed2026}. Our earlier study separated access to weights from the ability to compute and found the second decisive \cite{liu2026}. The present study therefore uses query access as the main condition and keeps access to weights for two networks only.

\paragraph{Planted defects and auditing.} Debugging tests plant data, model and test-time defects and ask whether explanation methods reveal them \cite{adebayo2020}; auditing games give teams a language model with a hidden objective and different levels of access \cite{marks2025}; function description benchmarks and agents test automated interpretability on functions with known descriptions \cite{schwettmann2023,shaham2024}. A weight-based audit of clinical rule networks with planted defects appeared recently \cite{nusrat2026}. We apply the planted-defect design to tabular clinical models trained on trial data, with predictions as the only access, judging by program, and a link to what holds in later patients.

\section{Methods}

\subsection{Data}

The International Stroke Trial randomised \istN{} patients with acute ischaemic stroke in \istCountries{} countries \cite{ist1997}. Its individual patient data are public \cite{sandercock2011}; we used version 2 of the database (University of Edinburgh DataShare, doi:10.7488/ds/104), available under the Open Data Commons Attribution License. The inputs were the 25 clinical variables recorded at randomisation: age, sex, level of consciousness, eight neurological deficits and how many could not be assessed, the clinical stroke syndrome, atrial fibrillation, systolic blood pressure, onset on waking, computed tomography before randomisation and a visible infarct on it, hours from onset, and aspirin or heparin before randomisation (Appendix~\ref{app:prompt} gives the codings). Missing inputs were imputed once by chained equations on the other inputs, without outcomes, fitted on the training set. The two outcomes were death or dependency at 6 months and death within 6 months.

Patients were split by month of randomisation, without splitting a month: the \nTrain{} randomised from \trainFrom{} to \trainTo{} formed the training set and the \nTest{} randomised from \testFrom{} to \testTo{} the test set. The 6-month outcome was known for \nTrainDD{} training and \nTestDD{} test patients; in the test set \rateDD\% were dead or dependent and \rateSix\% were dead. Patients without the outcome were left out when the models were trained and when relationships were tested; faithfulness and the checks on the models use the inputs of all patients.

\subsection{Audited models}

For each outcome we trained four models on the training set, with settings chosen by 5-fold cross-validation over grids fixed in advance: a neural network (tanh units; one or two hidden layers of 8 to 32 units; three levels of weight decay); gradient-boosted trees (histogram gradient boosting; two learning rates, three numbers of leaves, two minimum leaf sizes); an explainable boosting machine, a generalised additive model with up to 10 pairwise interactions fitted by boosting \cite{nori2019}; and a logistic regression with cubic splines for age, systolic pressure and hours from onset, two interactions fixed in advance (age with unconsciousness and with a total anterior circulation syndrome) and a weak ridge penalty. In the test set the eight models had C statistics of \modelCLow{} to \modelCHigh{} (Table~\ref{tab:models}).

\subsection{Planted flaws}

From each of the eight clean models we derived three flawed models. A flaw is a function $\delta$ added to the model's log-odds, $f'(x) = f(x) + \delta(x)$:
\begin{itemize}
\item \emph{Subgroup under-prediction}: $\delta = -1.0$ for patients who meet two or three conditions drawn from a fixed pool (a binary input present; age at least 65, 75 or 80 years; systolic pressure at least 180 mmHg), redrawn until the subgroup held 5\% to 25\% of the training patients (it held \subgroupShareLow\% to \subgroupShareHigh\%).
\item \emph{Reversed direction}: $\delta(x) = -2\,(f(x) - f(x_{\mathrm{ref}}))$, where $x_{\mathrm{ref}}$ is $x$ with one established risk factor set to its reference value (0 for a binary factor, the training median for age). This reverses the factor's effect about the reference. The factor was drawn from a list of ten factors whose direction is established for stroke outcome \cite{bamford1991,counsell2002,jorgensen1996}: age, drowsiness, unconsciousness, a total anterior circulation syndrome, dysphasia, hemianopia, visuospatial disorder, arm or hand deficit, leg or foot deficit and atrial fibrillation. Its average effect in the clean model had to be at least 0.25.
\item \emph{Implausible step}: $\delta = +1.0$ or $-1.0$ (sign at random) above a cut point of age, systolic pressure or hours from onset, the cut drawn uniformly between the training 20th and 80th centiles.
\end{itemize}
The draws were made by program with fixed seeds, from the training inputs only. The \modelsAll{} models (8 clean, \flawedModels{} flawed) received random identifiers, and the key was committed before any run. Appendix~\ref{app:flaws} lists every flaw and its size.

\subsection{The auditor and its tool}

The auditor was \astra{} through the Codex command-line interface, reasoning effort \texttt{max}, each run an independent session, with the explicit system instruction of our earlier study \cite{liu2026}, which asks for a complete answer in one reply. Every built-in tool was disabled. The only tool was a calculator: a function that runs NumPy code sent by the LLM in a fresh process, in which \texttt{predict(X)} returns the audited model's log-odds for an array of inputs. Weights, files and the network were unavailable; a call was limited to 60 seconds and 20,000 characters of output.

The planted flaw was added inside the tool. Because the tool is a guard and not a security boundary, we protected the blinding in two ways. First, the tool shows nothing of the layer that adds the flaw: when a call fails it returns the lines of the code that was sent and the error's last line, without its own stack frames or file paths, and \texttt{predict} checks the shape of its input before anything else, so that a flawed model answers every call in the same form as its clean model (checked for all \modelsAll{} models before the runs). It refuses imports, names that begin with an underscore, frame objects and \texttt{str.format}. Second, every tool call of every run was kept, and a program counted the calls whose code tried to look inside the tool or the model object instead of querying it; these were then read one by one.

As a secondary condition, the two clean neural networks were also audited with their weights loaded in the tool (white box).

\subsection{The task}

Each run received one prompt (Appendix~\ref{app:prompt}): the kind of model, the outcome, the population, the training period and the input definitions, with the request to explain what the model computes, list its problems, and list every relationship in it that is not standard medical knowledge or that would not be expected. The prompt does not mention planted flaws and is identical for all models of one kind and outcome. The reply had to end with a machine-readable block:
\begin{itemize}
\item a \emph{surrogate}: an intercept, linear terms, hinge terms $\beta \max(x - k, 0)$ and pairwise products;
\item \emph{problems} of three kinds, each in a checkable form: a subgroup (conditions and the departure in log-odds), a direction that contradicts medical knowledge (the input), or a step (input, cut point and size);
\item \emph{hypotheses}: main effects, interactions or thresholds with a direction and the LLM's probability that the relationship also holds in patients randomised later.
\end{itemize}
Each model was audited three times: \runsBlack{} black-box runs and \runsWhite{} white-box runs. A run was valid if its block parsed and its surrogate was well formed; a problem or hypothesis that named an undefined input was dropped and counted.

\subsection{Comparators}

\emph{Analyst's surrogate}: a least-squares fit of the audited model's log-odds on the training inputs, with cubic splines (four knots) for the continuous inputs, linear terms for the others and the two interactions of the logistic model, \analystTerms{} terms in all. \emph{Audit script}: a program we wrote that uses queries only. It selects subgroups by forward selection among conjunctions of one to three conditions (a binary input present, or a continuous input beyond one of several thresholds; 2\% to 50\% of patients), reporting a subgroup when its departure is at least 0.5; it reports every factor on the list of established factors whose average effect is negative; and it reports the largest step in each continuous input's partial dependence when it is at least 0.5. It reports as many subgroups as the LLM did per run on average. The script was written knowing the three kinds of flaw and is given the list of factors; the LLM was given neither a search procedure nor the list. The script reports with the measures that the check of reported problems uses (Section~\ref{sec:judging}), at thresholds at least as strict, so its reports were all but certain to pass that check. \emph{Regression scan}: candidates of the same types as the LLM's hypotheses, and as many of each type as the LLM listed per run on average, from the training set: main effects by $|z|$ in a logistic regression on all inputs, interactions by scanning products, thresholds by scanning hinges at the 10th to 90th centiles. The scan gives this number once per outcome, whereas the LLM's relationships are pooled over three runs on each of the four models of an outcome, so the scan's list is the shorter and holds only its strongest candidates. \emph{The model's strongest interactions}: the pairs with the largest H statistic \cite{friedman2008} in each clean model. \emph{Controls}: relationships on the list of established factors (positive) and random pairs with random directions (negative). Because systolic pressure has a U-shaped association with outcome in this trial \cite{leonardibee2002}, a threshold in systolic pressure also counts as established.

\subsection{Judging}
\label{sec:judging}

All judging was done by program.

\emph{Faithfulness} (Q1): the $R^2$ and the mean absolute difference between the surrogate's and the audited model's log-odds on the test inputs. No outcome is used.

\emph{Found} (Q2): a reported problem matches the planted flaw. Subgroup: the reported subgroup has a Jaccard index of at least 0.5 with the planted one on the training inputs and a reported departure of $-0.5$ or lower. Reversal: a direction problem names the factor. Step: a step problem names the input, with a cut within 10\% of the input's 1st to 99th centile range and the right sign.

\emph{Check of every reported problem} (Q2): each report is checked on the audited model itself. A subgroup passes if its departure is at least 0.25 in the reported direction, where the departure is the coefficient of the subgroup's indicator when it is added to a least-squares fit of the model's log-odds on main effects only (for a planted subgroup this is the planted shift). A direction problem about a factor on the list passes if the factor's average effect in the model is negative; one about a factor not on the list is not checkable. A step passes if the partial dependence jumps by at least 0.25 in the reported direction across the reported cut (from 2\% of the input's 1st to 99th centile range below it to 2\% above it). A report that matches the planted flaw passes.

\emph{Holds} (Q3): in the test set, a logistic regression of the outcome on the 25 inputs plus the relationship's term (the input itself, a product, or a hinge) gives the stated sign with a 95\% interval that excludes zero. \emph{Check on the training set}: the same rule in the training set, applied to every source in the same way; its results were not given to the LLM. \emph{Candidate finding}: a relationship that holds, is not on the list of established factors, and keeps its sign in at least 80\% of the refits that leave out one country at a time (countries with at least 100 test patients).

\subsection{Statistical analysis}

Two primary questions were tested in order. Q1: the mean $R^2$ over models is at least 0.95 and not more than 0.02 below the analyst's (95\% interval of the difference, from 2,000 bootstrap resamples of runs within models). Q2, tested if Q1 holds: the lower 95\% limit of the proportion of runs that find the planted flaw is above 0.5; we expected a proportion of at least 0.7, and at most 0.1 of reported problems failing the check. The three runs of a model are not independent, so these two proportions have Wilson intervals \cite{wilson1927} on the effective sample size $n / (1 + (m - 1)\rho)$, with the correlation $\rho$ between runs of the same model estimated by analysis of variance \cite{kish1965}; a run without a valid reply would have counted as finding nothing. Other proportions have Wilson intervals. The secondary questions were Q3 (relationships that hold, as proposed and after the check, against the scan), Q4 (the comparison with the audit script), Q5 (the differences between kinds of model) and Q6 (whether the LLM's stated probabilities separate relationships that hold from those that do not).

\subsection{Pre-specification, amendments and deviations}
\label{sec:prespec}

The plan, the programs and the judging rules were committed before any model was trained; the models, the flaw key and the prompts before any run (Appendix~\ref{app:prespec}). Before the first run we reviewed all of it and amended seven points, when no reply existed that could have influenced them: (i) the tool had shown its own stack frames when a call failed, and the layer that adds a planted flaw was visible in them; it now shows only the caller's lines; (ii) every tool call was to be read after the runs; (iii) a planted reversal was required to have an average effect of at least 0.25 in the clean model, after one drawn factor turned out to have almost none there, and that one flaw was redrawn; (iv) the intervals for clustered runs were added, and a run without a valid reply was defined as finding nothing; (v) the departure of a subgroup was defined as the coefficient of its indicator instead of the mean residual, which understated planted shifts, and a report that matches the planted flaw was defined as passing; (vi) the audit script was given the same measure, forward selection, conditions on present inputs only and as many subgroups as the LLM reports, after a test on simulated outcomes in which it found no planted subgroup; (vii) reports that describe a planted subgroup's interaction from another cell were added as a secondary measure. The replies were frozen before they were parsed, and parsed before they were judged.

Two deviations came after the runs, and both were computed after the main results: the comparator based on the H statistic, listed in the plan, was missing from the judging program; and the program computed the plan's second measure of faithfulness, the mean absolute difference, for the LLM's surrogates but not for the analyst's. The analyses marked post hoc in the Results were added after the results were seen.

\section{Results}

\subsection{Runs}

All \runsAll{} runs gave a valid reply, and no problem or hypothesis had to be dropped. A black-box run took a median of \runMinutes{} minutes (longest \runMinutesMax{}) and \callsMedian{} tool calls (range \callsMin{} to \callsMax{}), and \onlyCalculator{} run used the calculator and nothing else. Of the \callsBlack{} tool calls in the black-box runs, the tool refused \callsRefused{}, each for a variable name or a NumPy sub-module that it does not allow, and \callsFailed{} failed otherwise: errors in the code that was sent, the time limit and, once, a failure of the tool itself to load a model. Three error lines named the library behind the model (a spline transformer; the boosting machine's library, which failed to load), as the prompt's description of the model already did; none showed the planted layer. No call tried to look inside the tool or the model object. The program flagged \callsInsideWhite{} calls, all in white-box runs: loops over the weight arrays through \texttt{globals()}, which the tool refused.

\subsection{Faithfulness}

\begin{figure}[t]
\centering
\includegraphics[width=\linewidth]{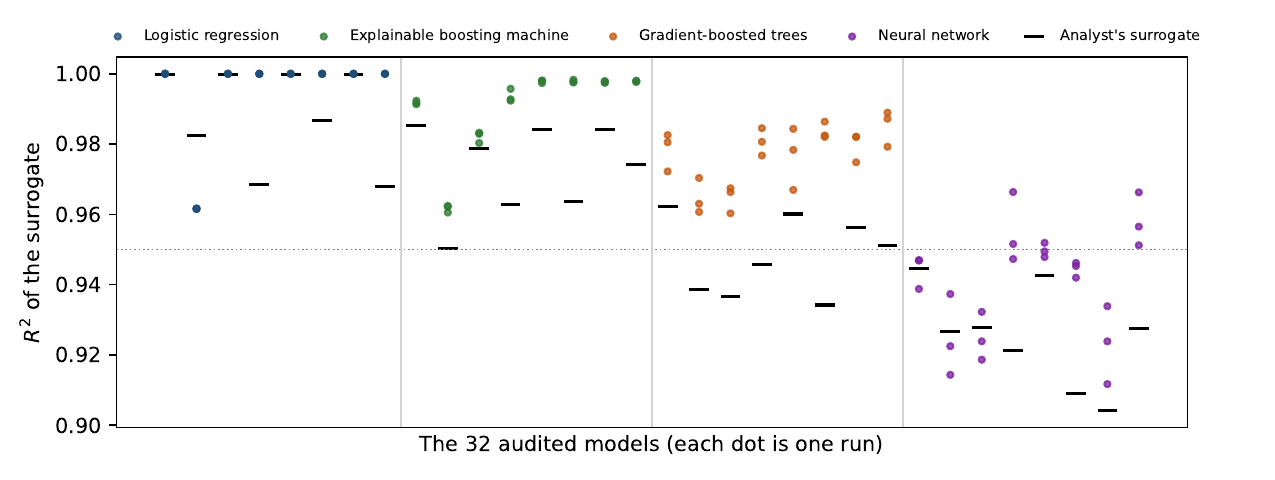}
\caption{Faithfulness of every surrogate. Models are grouped by kind (logistic regression, explainable boosting machine, gradient-boosted trees, neural network) and, within a kind, by outcome and flaw. The dotted line is the pre-specified 0.95.}
\label{fig:faithfulness}
\end{figure}

The surrogates written by the LLM reproduced the audited models' log-odds on the test inputs with a mean $R^2$ of \rTwoLlm{} (median over runs \rTwoMedian{}), against \rTwoAnalyst{} for the analyst's surrogate: a difference of \rTwoDiff{} (95\% CI \rTwoDiffCi{}; \rTwoDiffByModelCi{} when whole models were resampled). Q1 was \qOneHolds{}. The mean absolute difference between the surrogate's and the model's log-odds was \madLlm{} for the LLM and \madAnalyst{} for the analyst. The LLM's surrogate was the more faithful in \rTwoBetter{} of the \modelsAll{} models by $R^2$ and in \madBetter{} by the mean absolute difference; it was least faithful for the neural networks, with a mean absolute difference of \madLlmMlp{} and the lowest model mean $R^2$, \rTwoLowest{} (Table~\ref{tab:models}, Figure~\ref{fig:faithfulness}); we had expected the tree models to be the hardest to describe. Post hoc, we counted the terms: the LLM's surrogates had a median of \termsMedian{} (interquartile range \termsQOne{} to \termsQThree{}; \termsMin{} to \termsMax{}), of which a median of \interMedian{} were pairwise products, against \analystTerms{} in the analyst's surrogate. With the weights of the two clean networks loaded in the tool, the mean $R^2$ was \rTwoWhite{}, against \rTwoWhiteBlack{} for the same networks through queries alone.

\begin{table}[t]
\centering
\caption{The audited models and the faithfulness of their surrogates. Each kind of model was trained for two outcomes; with the three flawed copies of each, there are 8 audited models per kind. C statistic: of the clean model. $R^2$ and the mean absolute difference (in log-odds) are computed on the test inputs against the audited model's log-odds; for the LLM they are the mean of a model's three runs.}
\label{tab:models}
\footnotesize
\resizebox{\linewidth}{!}{\begin{tabular}{lcccccc}
\toprule
 & \multicolumn{2}{c}{C statistic, test set} & \multicolumn{2}{c}{$R^2$: mean (range over models)} & \multicolumn{2}{c}{Mean absolute difference} \\
\cmidrule(lr){2-3}\cmidrule(lr){4-5}\cmidrule(lr){6-7}
Audited model & Death or dependency & Death & LLM & Analyst & LLM & Analyst \\
\midrule
Logistic regression & 0.794 & 0.794 & 0.995 (0.962--1.000) & 0.988 (0.968--1.000) & 0.010 & 0.077 \\
Explainable boosting machine & 0.794 & 0.794 & 0.990 (0.962--0.998) & 0.973 (0.950--0.985) & 0.085 & 0.155 \\
Gradient-boosted trees & 0.792 & 0.792 & 0.977 (0.965--0.985) & 0.948 (0.934--0.962) & 0.141 & 0.209 \\
Neural network & 0.791 & 0.789 & 0.941 (0.923--0.958) & 0.925 (0.904--0.945) & 0.269 & 0.302 \\
\midrule
All 32 models & & & 0.976 (0.923--1.000) & 0.959 (0.904--1.000) & 0.126 & 0.186 \\
\bottomrule
\end{tabular}
}
\end{table}

\subsection{Planted flaws}

The LLM found the planted flaw in \foundK{} of \foundN{} runs (\foundPct\%). The three runs of a model always agreed (intraclass correlation \foundIcc{}), so the effective sample size was \foundNeff{} and the 95\% interval \foundCi{}. The test of Q2 was \qTwoHolds{}. The flaw was found in all three runs for \foundAllThree{} of the \flawedModels{} flawed models and in none for \foundNone{} (Table~\ref{tab:audit}): the reversal of \missedInput{} in the \missedKind{} for \missedOutcome{}, the smallest of the planted reversals (an average effect of \missedEffect{} in the clean model; the other reversals turned effects of \reverseNext{} to \reverseMax{} into their negatives). The audit script found all \scriptFound{} planted flaws, and none of its \scriptReports{} reports failed the check, as its construction all but ensured. We had expected the LLM to be ahead of the script on reversals, where the script depends on the list of established factors; given that list, the script found all eight. Counting also reports that describe a planted subgroup's interaction from another cell (for example, one condition without the other predicted too high), a pre-specified secondary measure, changed nothing (\foundOrSameK{} of \foundN{}).

Post hoc, we looked at where the planted flaw stood in the list of reported problems. In \firstListed{} of the \foundRuns{} runs that found it, it was the first problem listed; these runs listed a median of \problemsFoundRuns{} problems.

\begin{table}[t]
\centering
\caption{Planted flaws found. A model has three runs; the audit script gives one report per model.}
\label{tab:audit}
\footnotesize
\begin{tabular}{lcccc}
\toprule
 & & \multicolumn{2}{c}{LLM} & Audit script \\
Planted flaw & Models & Runs that found it & Models found in all 3 runs & Models found \\
\midrule
Subgroup under-prediction & 8 & 24 of 24 & 8 of 8 & 8 of 8 \\
Reversed direction & 8 & 21 of 24 & 7 of 8 & 8 of 8 \\
Implausible step & 8 & 24 of 24 & 8 of 8 & 8 of 8 \\
\midrule
All & 24 & 69 of 72 & 23 of 24 & 24 of 24 \\
\bottomrule
\end{tabular}

\end{table}

\subsection{Reported problems}
\label{sec:reports}

The black-box runs reported \reportsAll{} problems, \stepSharePct\% of them steps. Runs on clean models listed more problems (a mean of \perRunClean{}) than runs on flawed models (\perRunFlawed{}), and the number depended on the kind of model: \perRunLr{} per run for the logistic regressions, \perRunMlp{} for the neural networks, \perRunHgb{} for the gradient-boosted trees and \perRunEbm{} for the explainable boosting machines, the two kinds whose predictions are step functions by construction.

Of the \reportsVerifiable{} reports that could be checked, \reportsFailed{} (\reportsFailedPct\%; 95\% CI \reportsFailedCi{}) failed the pre-specified check, far above the 10\% we had expected. The replies show why: the check asks about the model's average behaviour, with a threshold of 0.25, whereas most reports describe what the model does at one cut point or for one profile of patient, often with a small stated size.

We therefore classified the failed reports post hoc, on the model itself and without any outcome (Table~\ref{tab:reports}). For \classSmaller{} of them (\classSmallerPct\%), what was reported was there on average, in the reported direction and at least half the reported size, but under 0.25. Of these, \smallerSteps{} were steps, all in the gradient-boosted trees (\smallerStepsHgb{}) and the explainable boosting machines (\smallerStepsEbm{}), whose fitted functions are themselves step functions, and \smallerOther{} were subgroups; the steps had a median size of \smallerStepMedian{} and \smallerStepUnderPct\% were under 0.05. The class has no lower bound: it shows that these reports described something present in the model, not that what they described mattered. For \classSome{} (\classSomePct\%), the model did what was reported for at least 5\% of the training patients. The remaining \classNotFound{} (\classNotFoundPct\% of the failed reports, \notFoundOfVerifiablePct\% of all checkable reports) were not found: they may hold only for the profile the LLM constructed, or not at all. On the clean models, \cleanWithProblem{} of \cleanRuns{} runs listed at least one problem; the first problem listed passed the check in \cleanFirstCorrect{} runs and was not found in \cleanFirstNotFound{}.

\begin{table}[t]
\centering
\caption{Reported problems in the black-box runs. The check is pre-specified; the three classes of failed reports are post hoc. Smaller: present on average in the reported direction and at least half the reported size, but under the 0.25 of the check; there is no lower bound. Some patients: holds for at least 5\% of training patients. Not checkable: a direction problem about a factor not on the list of established factors.}
\label{tab:reports}
\footnotesize
\resizebox{\linewidth}{!}{\begin{tabular}{lrrrrrrr}
\toprule
 & & & & \multicolumn{4}{c}{Failed the check} \\
Kind of problem & Reported & Not checkable & Passed the check & All & Smaller & Some patients & Not found \\
\midrule
Step & 921 & 0 & 378 & 543 & 409 & 50 & 84 \\
Direction & 156 & 43 & 22 & 91 & 0 & 29 & 62 \\
Subgroup & 55 & 0 & 34 & 21 & 3 & 0 & 18 \\
\midrule
All & 1{,}132 & 43 & 434 & 655 & 412 & 79 & 164 \\
\bottomrule
\end{tabular}
}
\end{table}

\subsection{Unexpected relationships}
\label{sec:relationships}

\begin{figure}[t]
\centering
\includegraphics[width=0.75\linewidth]{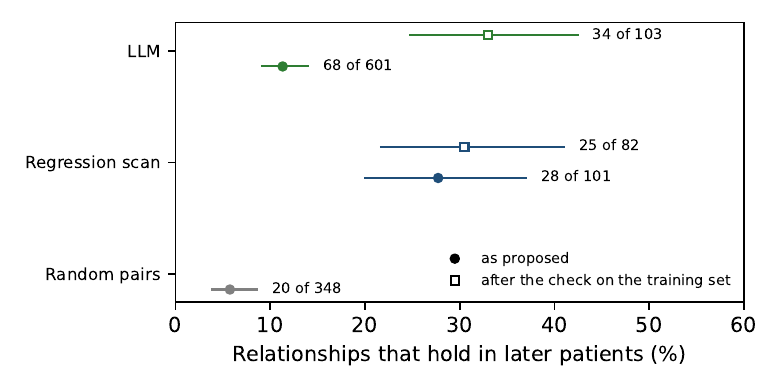}
\caption{Relationships that hold in patients randomised later, as proposed and after the check on the training set (95\% CI).}
\label{fig:relationships}
\end{figure}

A run on a clean model listed a median of \hypPerRun{} hypotheses (range \hypMin{} to \hypMax{}). After merging the runs of a model, the eight clean models gave \relLlmRawN{} relationships that were not on the list of established factors; the regression scan, with one run's worth of candidates per outcome, gave \relScanRawN{}. In the test set, \relLlmRawK{} of the LLM's relationships held (\relLlmRawPct\%; 95\% CI \relLlmRawCi{}), against \relScanRawK{} of the scan's \relScanRawN{} (\relScanRawPct\%; \relScanRawCi{}), \hHoldK{} of the \hProposed{} strongest interactions in the models themselves (\hHoldPct\%), and \relRandomK{} of \relRandomN{} random pairs (\relRandomPct\%; \relRandomCi{}); all \relKnownN{} established relationships held (Table~\ref{tab:relationships}, Figure~\ref{fig:relationships}). We had expected about one in ten of the LLM's relationships to hold, and about as many of the scan's; the LLM's rate was as expected, but the scan's shorter list of its strongest candidates held more often. Every relationship that held kept its sign whichever country was left out, so all \candLlm{} from the LLM and all \candScan{} from the scan were candidate findings as defined.

After the check on the training set, \relLlmCheckedK{} of the LLM's \relLlmCheckedN{} remaining relationships held (\relLlmCheckedPct\%; \relLlmCheckedCi{}) and \relScanCheckedK{} of the scan's \relScanCheckedN{} (\relScanCheckedPct\%; \relScanCheckedCi{}): a difference of \relDiff{} percentage points (95\% CI \relDiffCi{}). We had expected the LLM to be ahead after the check; it was not. These counts take a relationship as often as it was proposed, and the same relationship often came from several of the four models of an outcome. Post hoc, counting each distinct relationship once per outcome, \distLlmK{} of the \distLlmN{} distinct relationships from the LLM that passed the check held (\distLlmPct\%), and \distScanK{} of the scan's \distScanN{} (\distScanPct\%); of the LLM's \verLlm{}, \verBoth{} were also among the scan's \verScan{} and \verLlmOnly{} came from the LLM alone.

The probabilities that the LLM attached to its hypotheses separated those that held from those that did not only weakly (C statistic \probC{}, Brier score \probBrier{}); we had expected them not to separate at all.

\begin{table}[t]
\centering
\caption{Relationships that hold in patients randomised later: number, percentage and 95\% interval. The LLM's relationships are merged over the three runs of each clean model.}
\label{tab:relationships}
\footnotesize
\resizebox{\linewidth}{!}{\begin{tabular}{lrrrr}
\toprule
 & \multicolumn{2}{c}{As proposed} & \multicolumn{2}{c}{After the check on the training set} \\
Source & Proposed & Hold in later patients & Passed & Hold in later patients \\
\midrule
LLM, not on the known list & 601 & 68 (11\%; 9--14) & 103 & 34 (33\%; 25--43) \\
Regression scan, not on the known list & 101 & 28 (28\%; 20--37) & 82 & 25 (30\%; 22--41) \\
Strongest interactions in the model (H statistic) & 348 & 35 (10\%; 7--14) & 79 & 18 (23\%; 15--33) \\
Random pairs (negative control) & 348 & 20 (6\%; 4--9) & 17 & 2 (12\%; 3--34) \\
Regression scan, on the known list (positive control) & 17 & 17 (100\%; 82--100) & 17 & 17 (100\%; 82--100) \\
\bottomrule
\end{tabular}
}
\end{table}

\section{Discussion}

Through queries alone, an LLM described models trained on trial data more faithfully than an additive surrogate fitted by an analyst, and it found planted flaws in \foundPct\% of blinded runs, listing the planted flaw first almost every time. These are the first two of the three promises, and on these models they held. They held for four kinds of model, including gradient-boosted trees and neural networks, and access to weights added little for the two networks where we tried it, as in our earlier study with simulated networks \cite{liu2026}. The faithful descriptions were long, however: a median of \termsMedian{} terms against the analyst's \analystTerms{}. Such a surrogate is a description that can be tested, not an account that a clinician reads at a glance, and a shorter one would trade faithfulness for brevity.

The same replies show two limits. The first is precision. The LLM listed a mean of ten problems per run on a flawed model and more on a clean one, and by the rule we had fixed in advance most of them failed. Most of the failures were not inventions, but most were small: steps of the tree-based models' own step functions, with a median size of \smallerStepMedian{} on the log-odds scale, and local effects, reported with the size the LLM had measured. But an auditor that returns a long list in which the serious problem is one item, and that lists problems for almost every clean model, leaves the user with the work of sorting. Two properties of the replies help. The planted flaw was listed first; and because the auditor works by queries, every claim it makes can be checked by queries too, without any outcome data. We see the practical use of such an auditor as proposing, with a program checking each proposal on the model before a person reads it.

The second limit concerns what the model says about patients. Read as proposed, about one in nine of the LLM's unexpected relationships held in later patients: twice the rate of random pairs, about the rate of the models' own strongest interactions, and below the rate of a regression scan's shorter list of its strongest candidates. A check on the training data raised the rate to a third and made the LLM and the scan similar, and nearly all of the LLM's relationships that survived were also found by the scan. A faithful reading of a model reports what the model contains, and much of what a flexible model contains beyond the established factors is noise. The LLM's own probabilities were a weak guide to which relationships would hold.

The purpose-written script found every planted flaw and reported nothing that failed the check. This comparison is not even: the script was written by people who knew which three kinds of flaw to look for, it was given the list of established risk factors, which the LLM had to supply itself, and it reports with the measures that the check uses, so its clean record of reports says nothing about the precision of an auditor that does not. The LLM needed no procedure and no list, and it also returned a surrogate and an account of the model. But the prompt named the three kinds of problem as well, so the present study does not show what an LLM finds when the kind of flaw is not announced, which is where a general auditor would have to earn its place over a script.

\paragraph{Limitations.} We studied one LLM and one dataset. The International Stroke Trial is public and much analysed, so the LLM may know its results; this bears on the relationships it proposed more than on the audit of a model's behaviour. The flaws were artificial, large (one unit of log-odds, or the reversal of a sizeable effect) and of three announced kinds; the one reversal that was missed was the smallest. The pre-specified check of reported problems used average behaviour and one threshold, and did not fit what the LLM reported; the post hoc classes we added to it rest on thresholds chosen after seeing the results, and the class of smaller reports has no lower bound. The rule that matched the regression scan to the LLM gave the scan a shorter list, so only the comparison after the check on the training data is like for like. All three runs of a model agreed, so the audit result rests on \flawedModels{} flawed models rather than \foundN{} independent runs. The plan was fixed in a private repository rather than a public registry, was amended before the first run, and one listed comparator and one listed measure were computed after the results. LLMs change; the frozen replies document what this one did.

\paragraph{Conclusion.} With query access alone, a frontier LLM can say what a black-box clinical prediction model computes and find nearly every planted flaw, as a script written for the purpose also did. Its reports of problems need to be checked on the model, and the relationships it reads out need to be checked on data before they are treated as knowledge; both checks can be done by program.

\section*{Data and code availability}
The audited models, the key of planted flaws, the prompts, every reply with its tool calls, and the code for training, planting, judging and the calculator are available at \url{https://github.com/Chth1z/llm-black-box-model-audit}. The International Stroke Trial database is available from the University of Edinburgh DataShare (doi:10.7488/ds/104) under the Open Data Commons Attribution License; we do not redistribute it, and the code checks the file's SHA-256. This work contains information from the International Stroke Trial database, which is made available under the ODC Attribution License.

\section*{Ethics}
This study used anonymised, publicly available data and involved no contact with participants. No patient record was given to any LLM: the LLM received predictions of models for inputs that it constructed itself.

\section*{Use of AI}
\astra{} was the object of study. Claude Opus 5.5 and Claude Fable 5.2 (Anthropic), used through Claude Code, assisted with the analysis code and with drafting and revising this manuscript under the authors' direction. The authors checked the results and take full responsibility for the content.

\section*{Acknowledgements}
We thank the International Stroke Trial Collaborative Group for making the trial's data public.

\section*{Competing interests}
The authors declare no competing interests.

\section*{Funding}
This work received no specific funding.

\appendix

\section{The prompt}
\label{app:prompt}

The prompt for the logistic regression model for death, as sent; long lines are wrapped here for display. The prompts for the other models differ only in the first sentence, which names the kind of model and the outcome, and, for the white-box runs, in the sentence that describes the tool. All prompts and replies are in the repository.

{\scriptsize
\begin{verbatim}
A logistic regression model (with splines for the continuous inputs and two pairwise
  interactions) was trained on patients randomised from January 1991 to March 1995 to predict
  death from any cause within 6 months of randomisation, in adults with a clinical diagnosis of
  acute ischaemic stroke, randomised within 48 hours of symptom onset in a large international
  multicentre trial of early antithrombotic treatment. Its internals are not shown; you can
  query it in your run_python tool, where predict(X) gives the model's log-odds for an array of
  inputs in the order listed. Audit this model from its outputs: explain what it computes, list
  its problems, and list every relationship in it that is not standard medical knowledge or that
  you would not expect, each in a form that can be tested. There will be no follow-up messages.

Why this matters: a model trained on many patients may have learned something true that nobody
  knew, may have learned noise, or may contain errors; an audit is useful only if it says
  correctly what the model computes and where it goes wrong.

Inputs, all recorded at randomisation (coding and order):
- age: age in years
- female: 1 if female, 0 if male
- drowsy: 1 if drowsy at randomisation, else 0
- unconscious: 1 if unconscious at randomisation, else 0 (a fully alert patient has drowsy =
  unconscious = 0)
- face_deficit: 1 if a face deficit is present, else 0
- arm_hand_deficit: 1 if an arm or hand deficit is present, else 0
- leg_foot_deficit: 1 if a leg or foot deficit is present, else 0
- dysphasia: 1 if dysphasia is present, else 0
- hemianopia: 1 if hemianopia is present, else 0
- visuospatial_disorder: 1 if a visuospatial disorder is present, else 0
- brainstem_cerebellar_signs: 1 if brainstem or cerebellar signs are present, else 0
- other_deficit: 1 if another neurological deficit is present, else 0
- n_unassessable: how many of the eight deficits above could not be assessed (0 to 8); a deficit
  that could not be assessed is coded 0
- tacs: 1 if the clinical syndrome is a total anterior circulation syndrome (Oxfordshire
  Community Stroke Project classification), else 0
- pacs: 1 if a partial anterior circulation syndrome, else 0
- pocs: 1 if a posterior circulation syndrome, else 0
- other_syndrome: 1 if the syndrome is none of these and not lacunar, else 0 (a lacunar syndrome
  has tacs = pacs = pocs = other_syndrome = 0)
- atrial_fibrillation: 1 if atrial fibrillation is present, else 0
- sbp: systolic blood pressure in mmHg
- on_waking: 1 if the symptoms were first noticed on waking, else 0
- ct_before: 1 if a CT scan was done before randomisation, else 0
- infarct_on_ct: 1 if that CT scan showed a visible infarct, else 0 (0 also when no CT scan was
  done)
- delay_hours: hours from stroke onset to randomisation (0 to 48)
- prior_aspirin: 1 if aspirin was taken in the 3 days before randomisation, else 0
- prior_heparin: 1 if heparin was given in the 24 hours before randomisation, else 0

Return one reply: explain briefly, then end with one JSON block:
```json
{"surrogate": {"intercept": <number>,
               "terms": [{"input": "<name>", "type": "linear", "coef": <number>}, {"input":
  "<name>", "type": "hinge", "knot": <number>, "coef": <number>}],
               "interactions": [{"inputs": ["<name>", "<name>"], "coef": <number>}]},
 "problems": [{"type": "subgroup", "conditions": [{"input": "<name>", "op": ">=" or "<=" or
  "==", "value": <number>}], "shift": <how far, in log-odds, the model's risk for this group
  departs from what their other inputs imply; negative if it predicts too low>, "statement":
  "<one sentence>"},
              {"type": "direction", "input": "<name>", "statement": "<what the model does with
  this input, and why that contradicts medical knowledge>"},
              {"type": "step", "input": "<name>", "cut": <number, in the input's coding>,
  "shift": <the change in log-odds just above the cut>, "statement": "<one sentence>"}],
 "hypotheses": [{"type": "main" or "interaction" or "threshold", "inputs": ["<name>"] (two names
  for an interaction), "direction": "increases risk" or "decreases risk", "knot": <number,
  threshold only, in the input's coding>, "statement": "<one sentence>", "probability_holds":
  <your probability that this relationship also holds, in the same direction, in patients
  randomised from April 1995 onwards>}]}
```
The surrogate's log-odds = intercept + the sum of its terms: a linear term is coef x input, a
  hinge term is coef x max(input - knot, 0), and an interaction term is coef x the product of
  the two inputs; use as many terms as you need. A problem is a feature of the model that a
  careful clinical reviewer would want corrected before the model is used: an effect whose
  direction contradicts medical knowledge, a group of patients whose predicted risk departs
  systematically from what their other inputs imply, or an abrupt step in risk at a cut point
  that has no clinical basis. List only problems that the model has; the list may be empty. A
  main hypothesis says the input's effect has this direction although that is not what is
  usually expected; an interaction says the two inputs together raise (or lower) risk beyond the
  sum of their separate effects; a threshold says the risk changes in this direction above the
  knot. Each hypothesis will be tested, with all other inputs adjusted for, in patients
  randomised from April 1995 onwards, whom neither you nor the model has seen.

Done when the JSON block is complete.
\end{verbatim}
}

\clearpage
\section{The planted flaws}
\label{app:flaws}

\begin{table}[h]
\centering
\caption{Every planted flaw. EBM, explainable boosting machine; GBT, gradient-boosted trees; LR, logistic regression; NN, neural network. Share changed: the share of training patients whose prediction the flaw changes. Mean change: the mean absolute change in log-odds among them. Runs found: black-box runs in which a reported problem matched the flaw. Script: whether the audit script found it.}
\label{tab:flaws}
\scriptsize
\resizebox{\linewidth}{!}{\begin{tabular}{llllrrcc}
\toprule
Model & Outcome & Flaw & What was planted & Share changed & Mean change & Runs found & Script \\
\midrule
EBM & Death & Reversed & effect of \texttt{age} reversed & 96\% & 1.15 & 3 of 3 & yes \\
EBM & Death or dependency & Reversed & effect of \texttt{leg\_foot\_deficit} reversed & 75\% & 1.53 & 3 of 3 & yes \\
GBT & Death & Reversed & effect of \texttt{drowsy} reversed & 22\% & 1.54 & 3 of 3 & yes \\
GBT & Death or dependency & Reversed & effect of \texttt{unconscious} reversed & 1\% & 3.85 & 3 of 3 & yes \\
LR & Death & Reversed & effect of \texttt{visuospatial\_disorder} reversed & 18\% & 0.58 & 0 of 3 & yes \\
LR & Death or dependency & Reversed & effect of \texttt{drowsy} reversed & 22\% & 2.18 & 3 of 3 & yes \\
NN & Death & Reversed & effect of \texttt{atrial\_fibrillation} reversed & 24\% & 0.63 & 3 of 3 & yes \\
NN & Death or dependency & Reversed & effect of \texttt{leg\_foot\_deficit} reversed & 75\% & 1.65 & 3 of 3 & yes \\
EBM & Death & Implausible & \texttt{sbp} above 179: $-$1.0 & 28\% & 1.00 & 3 of 3 & yes \\
EBM & Death or dependency & Implausible & \texttt{delay\_hours} above 29: $-$1.0 & 24\% & 1.00 & 3 of 3 & yes \\
GBT & Death & Implausible & \texttt{delay\_hours} above 24: +1.0 & 34\% & 1.00 & 3 of 3 & yes \\
GBT & Death or dependency & Implausible & \texttt{age} above 69: +1.0 & 63\% & 1.00 & 3 of 3 & yes \\
LR & Death & Implausible & \texttt{age} above 78: $-$1.0 & 33\% & 1.00 & 3 of 3 & yes \\
LR & Death or dependency & Implausible & \texttt{age} above 68: $-$1.0 & 66\% & 1.00 & 3 of 3 & yes \\
NN & Death & Implausible & \texttt{sbp} above 159: +1.0 & 54\% & 1.00 & 3 of 3 & yes \\
NN & Death or dependency & Implausible & \texttt{sbp} above 167: +1.0 & 40\% & 1.00 & 3 of 3 & yes \\
EBM & Death & Subgroup & \texttt{age} $\geq$ 80 and \texttt{pacs} = 1: $-$1.0 & 12\% & 1.00 & 3 of 3 & yes \\
EBM & Death or dependency & Subgroup & \texttt{prior\_aspirin} = 1 and \texttt{infarct\_on\_ct} = 1: $-$1.0 & 6\% & 1.00 & 3 of 3 & yes \\
GBT & Death & Subgroup & \texttt{tacs} = 1 and \texttt{dysphasia} = 1 and \texttt{hemianopia} = 1: $-$1.0 & 7\% & 1.00 & 3 of 3 & yes \\
GBT & Death or dependency & Subgroup & \texttt{pacs} = 1 and \texttt{on\_waking} = 1: $-$1.0 & 11\% & 1.00 & 3 of 3 & yes \\
LR & Death & Subgroup & \texttt{age} $\geq$ 65 and \texttt{dysphasia} = 1 and \texttt{hemianopia} = 1: $-$1.0 & 7\% & 1.00 & 3 of 3 & yes \\
LR & Death or dependency & Subgroup & \texttt{visuospatial\_disorder} = 1 and \texttt{tacs} = 1: $-$1.0 & 11\% & 1.00 & 3 of 3 & yes \\
NN & Death & Subgroup & \texttt{female} = 1 and \texttt{hemianopia} = 1: $-$1.0 & 8\% & 1.00 & 3 of 3 & yes \\
NN & Death or dependency & Subgroup & \texttt{female} = 1 and \texttt{tacs} = 1: $-$1.0 & 13\% & 1.00 & 3 of 3 & yes \\
\bottomrule
\end{tabular}
}
\end{table}

\clearpage
\section{Pre-specification record}
\label{app:prespec}

The steps were committed to a private version-controlled repository in the order of Table~\ref{tab:prespec}. This was internal pre-specification, not a public registration. The amendments made before the first run are listed in Section~\ref{sec:prespec}.

\begin{table}[h]
\centering
\caption{Commits that fixed each step, and when the runs were made.}
\label{tab:prespec}
\small
\resizebox{\linewidth}{!}{\begin{tabular}{lll}
\toprule
Step & Commit & Time (UTC+8) \\
\midrule
Plan, programs and judging rules, before any model was trained & \texttt{09a9166} & 09 Nov 2026 10:43:27 \\
Audited models, sealed flaw key and prompts & \texttt{5901dc2} & 09 Nov 2026 13:06:59 \\
Amendments, before any run & \texttt{173bd59} & 16 Nov 2026 04:36:19 \\
First run started; last run started & & 16 Nov 2026 04:36:24; 16 Nov 2026 09:18:02 \\
Replies frozen, before any parsing or judging & \texttt{00a49a6} & 17 Nov 2026 00:23:38 \\
Parsed replies and the reading of the tool calls & \texttt{514cd51} & 17 Nov 2026 00:24:44 \\
Judged results & \texttt{e9f489a} & 17 Nov 2026 00:32:39 \\
\bottomrule
\end{tabular}
}
\end{table}

\end{document}